\documentclass{article}

\usepackage[width=122mm,left=12mm,paperwidth=146mm,height=193mm,top=12mm,paperheight=217mm]{geometry}
\usepackage[utf8]{inputenc} % allow utf-8 input
\usepackage[T1]{fontenc}    % use 8-bit T1 fonts
\usepackage{hyperref}       % hyperlinks
\usepackage{url}            % simple URL typesetting
\usepackage{booktabs}       % professional-quality tables
\usepackage{amsfonts}       % blackboard math symbols
\usepackage{nicefrac}       % compact symbols for 1/2, etc.
\usepackage{microtype}      % microtypography
\usepackage{graphicx}
\usepackage{subfig}
\usepackage{amsmath}
\usepackage{makecell}
\usepackage{multirow}
\usepackage{algpseudocode}
\usepackage{algorithm}
\date{}
\usepackage{authblk}

\title{Comb Convolution for Efficient Convolutional Architecture}

\author{Dandan Li\textsuperscript{\rm 1}\hspace{0.2cm}  Yuan Zhou\textsuperscript{\rm 1}\hspace{0.2cm} Shuwei Huo\textsuperscript{\rm 1}\hspace{0.2cm} Sun-Yuan Kung\textsuperscript{\rm 2}\\ % All authors must be in the same font size and format. Use \Large and \textbf to achieve this result when breaking a line
\textsuperscript{\rm 1}Tianjin University
\hspace{1.3cm}\textsuperscript{\rm 2}Princeton University
\\ \text{\{lidandan95,zhouyuan,huosw\}@tju.edu.cn}
\\ \text{kung@princeton.edu}
 }

\begin{document}

\maketitle

\begin{abstract}
  Convolutional neural networks (CNNs) are inherently suffering from massively redundant computation (FLOPs) due to the dense connection pattern between feature maps and convolution kernels. Recent research has investigated the sparse relationship between channels, however, they ignored the spatial relationship within a channel. In this paper, we present a novel convolutional operator, namely comb convolution, to exploit the intra-channel sparse relationship among neurons. The proposed convolutional operator eliminates nearly $50\%$ of connections by inserting uniform mappings into standard convolutions and removing about half of spatial connections in convolutional layer. Notably, our work is orthogonal and complementary to existing methods that reduce channel-wise redundancy. Thus, it has great potential to further increase efficiency through integrating the comb convolution to existing architectures. Experimental results demonstrate that by simply replacing standard convolutions with comb convolutions on state-of-the-art CNN architectures ($e.g.$, VGGNets, Xception and SE-Net), we can achieve 50\% FLOPs reduction while still maintaining the accuracy.
\end{abstract}

\section{Introduction}

Deep learning methods, especially convolutional neural networks (CNNs), have revolutionized the field of computer vision. The central building block of CNNs is the convolution operator that enables networks to construct informative features by fusing both spatial and channel-wise information within local receptive fields. Such feature extraction ability of CNNs has proven great capability of solving various visual recognition tasks \cite{43}, including image classification, semantic segmentation, object detection, target tracking, action recognition , attention prediction, salient object detection. However, standard convolution operations inherently suffer from the crucial problem: the dense connectivity pattern between feature maps and convolution kernels forces networks to repeatedly process the same feature between adjacent locations, leading to substantial redundancy and high computation overhead. Thus, it is critical to design efficient convolution operators for convolutional architectures.

Recent advances in accelerating convolution have focused primarily on reducing channel-wise redundancy in convolutional feature maps. By constructing structured sparse relationship between channels, most prior works,  $e.g.$, ResNeXt \cite{11}, Xception \cite{12}, ShuffleNets \cite{13,14} and MobileNets \cite{15,16}, Deep roots \cite{45}, CondenseNet \cite{17} and IGCNets \cite{18,19,20}, leads to consistent efficiency gains in the structure of networks. However, massive redundancy also exists in the spatial dimension of the feature maps, which restrict the further improvement of network efficiency.

\begin{figure}
\centering
{\includegraphics[width=1\columnwidth]{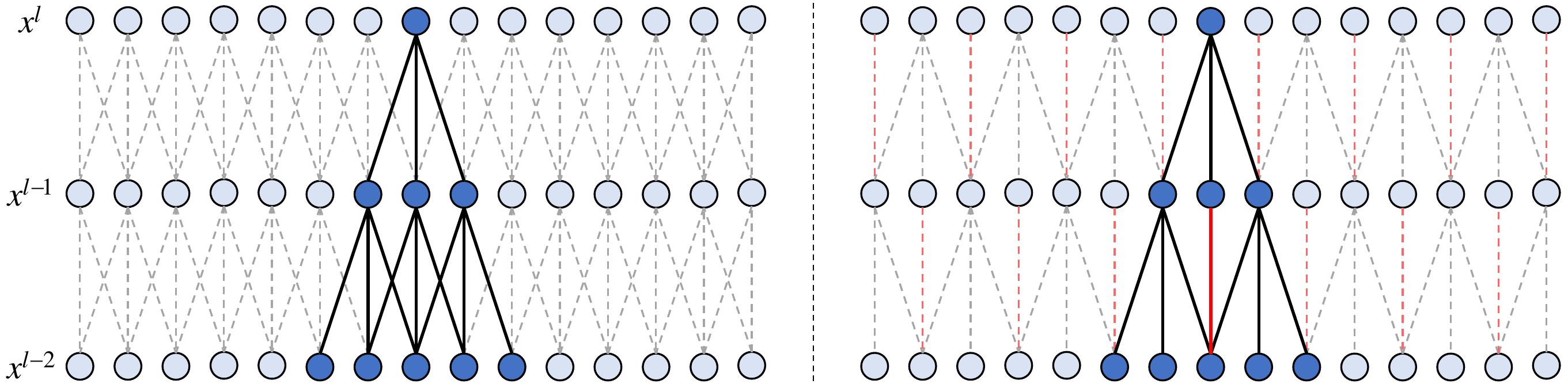}}\
\caption{Illustrations of a stack of standard convolutional layers($left$) and comb convolutional layers($right$). Different colored lines indicate different ways of mapping. The black line is convolution mapping and the red line is the uniform mapping. Two sets of locations are highlighted, corresponding to the highlighted units above. Stacked comb convolution enable networks to have the same receptive fields with just a few layers.} \label{fig1}
\end{figure}

Unlike most existing methods that eliminate channel dimensional redundancy, we focus on spatial dimension and present a novel convolutional operator, which eliminates nearly $50\%$ of interlayer connections and reduces computational complexity (FLOPs) by half. In our method, we insert uniform mappings into standard convolutions. As illustrated in Figure ~\ref{fig1}($right$), there are two kind of feature mappings, $i.e.$, convolution mapping and uniform mapping, which are both checkerboard symmetry in the spatial distribution. Since its connectivity pattern gives the appearance of a comb, we refer our approach as comb convolution.

As an alternative to standard convolution, comb convolution saves a significant amount of computation without sacrificing model capacity. Moreover, comb convolution, in a similar spirit as ResNet-like \cite{23,24,25,26} and DenseNet-like \cite{27,28,29} architectures, exposes a new kind of feature reuse, namely point-wise feature reuse. It helps strike an excellent trade-off between representation capability and computational cost, yielding models with high capacity and efficiency.

Comb convolution can be readily trained end-to-end by standard back-propagation and can easily replace their plain counterparts in existing CNNs, giving rise to efficient convolutional networks. Note that our work focuses on reducing spatial redundancy, which is orthogonal to existing methods that utilize channel-wise operations. Thus, it has great potential to further increase efficiency through integrating the comb convolution to existing architectures. Ablation studies compare the performances and costs under various configurations, which provides a useful guidance for network design. Extensive experimental results show that by simply replacing the standard convolutions with comb convolutions, the proposed approach can significantly boost speed of the state-of-the-art CNNs, $e.g.$, VGGNets\cite{21}, Xception and SE-Net\cite{22}.

\section{Comb Convolution}

In standard convolution operation, the previous layer’s feature maps $x^{l-1}$ are convolved with a distinct kernel $w^l_j$ and put through a linear or non-linear activation function to form one output feature map $x^l_j$. In general, we have:
\begin{equation}
x^l_j=\mathcal{F}(x^{l-1},w^l_j) \label{eq1}
\end{equation}
where $\mathcal{F}$ denotes the convolution function to be learned. Note that we did not add the bias term in all equations to keep the equation clean.

As a means to both eliminate spatial dimensional redundancy and maintain model representation capability, we apply a binary mask tensor of zeroes and ones to each layer. When applied to equation ~\ref{eq1}, we have:
\begin{equation}
x^l_j=Mask^l_j\cdot \mathcal{F}(x^{l-1},w^l_j)+(\textbf{1}-Mask^l_j)\cdot \mu(x^{l-1})
\end{equation}
where $Mask^{l}_j$ refers to the mask matrix corresponding to the output channel, \textbf{1} refers to an all-ones matrix. $\mu$ represents a uniform mapping operator, which can be represented as:
\begin{equation}
\mu(x^{l-1})=\frac{1}{C^l}\sum_i^{C^{l-1}}x^{l-1}_i
\end{equation}
where $C^l$ and $C^{l-1}$ are the total number of output channels and input channels, respectively. More specifically, for each location $(p,q)$ on the mask matrix $Mask^{l}_j$, we have:
\begin{equation}
Mask^l_j(p,q)=\left\{
\begin{array}{rcl}
1 & & {p+q=2k}\\
0 & & {p+q=2k+1}\\
\end{array} \right.\forall k\in \mathbb{N}
\end{equation}
where $(p,q)$ iterates over the output feature map $x^l_j$, the symbol $\mathbb{N}$ stands for the set of nonnegative integers. Note that $Mask^l_j$ is the same size as $x^l_j$. We also note that in the special case where both the input and output have only one channel, $i.e.$, $C^{l-1}=C^l=1$, uniform mapping becomes to identity mapping. The output is $x^l$ which can be formulated as:
\begin{equation}
x^l=Mask^l\cdot \mathcal{F}(x^{l-1},w^l)+(\textbf{1}-Mask^l)\cdot x^{l-1}
\end{equation}
Through embedding uniform mappings into convolutional layer, the feature mapping in the network has two types: convolution mapping and uniform mapping. When the sum of spatial coordinates of a unit is even, a unit will be obtained by convolution mapping; otherwise, obtained by the uniform (or identity) mapping.

We hypothesize that the learning capability of comb convolutional layers is a function of both the sparsity of the weight matrix as well as the complexity of the mapping. Based on this insight, we develop a fixed mapping permutation called mask interleaving, $i.e.$, the mask between adjacent channels are also both checkerboard symmetry in the spatial distribution. Hence, the units in intermediate layers, either with or without standard convolution operations have dependency in each other. Algorithm ~\ref{algorithmic1} shows the details of this procedure. Note that mask interleaving is differentiable, which means it can be plugged into neural networks for training by back propagation. Experimental results in Table ~\ref{table1} demonstrate the effectiveness of the mask interleaving architecture.

Furthermore, we also apply checkerboard symmetry mask tensors to the convolutional layer alternately. By stacking multiple layers, comb convolution can keep the same receptive field of output units as standard convolution, thereby contributes to efficiently model the dependencies in spatial dimensions. As illustrated in Figure ~\ref{fig1}, we highlight an output unit, and also highlight the input units that affect this unit. The receptive field of the output unit of a comb convolutional neural network is equal to the receptive field of the output unit of a standard convolutional network. Note that the effect increases when multiple comb convolution operators are stacked one after another. This is verified in experiments, as discussed in Section \ref{II}.

\begin{algorithm}[htb]
\caption{Comb convolution operator. $L$ is the network depth and $C^l$ is the number of current layer channels. $x^{l-1}_{slice}$ corresponds to the slice that generates the output unit.}
\label{algorithm1}
\begin{algorithmic}[1]
\Require
  The previous layer output $x^{l-1}$, and the weights $w^l_j, l\in{\{1,\dots,L\}}, j\in{\{1,\dots,C^l\}}$.
\Ensure
  The current layer output $x^l$.
\For{each output location $(p,q,j)$ iterations}
  \If {$(p+q+j)=0$}
    \State $x^l_j(p,q,j)\gets \mathcal{F}(x^{l-1}_{slice},w^l_j)$
  \Else
    \State $x^l_j(p,q,j)\gets \mu(x^{l-1}(p,q))$
  \EndIf
  \State $x^l\gets x^l_j$
\EndFor\\
\Return $x^l$
\end{algorithmic}
\label{algorithmic1}
\end{algorithm}

\section{Understanding Comb Convolution}
\subsection{Computational efficiency}

Comb convolution inserts uniform mappings into the convolution, thereby reduces the computation by half. Suppose a 2D convolutional operation at a certain layer takes an input tensor $x^l \in \mathbb{R}^{M\times M\times C^{l-1}}$, and applies convolutional kernel $w^{l} \in \mathbb{R}^{C^{l-1}\times K\times K\times C^l}$ using unit stride to produce an output tensor $x^l \in \mathbb{R}^{N\times N\times C^l}$. In this case, the standard convolution operator has a computational cost of $N\times N\times K\times K\times C^{l-1}\times C^l$. In contrast, due to the removal of connections (the amount is about $\frac{1}{2}\times (K\times K-1)\times N\times N\times C^{l-1}\times C^l$), comb convolution operator has the computational cost of $(\frac{1}{2}\times K\times K\times C^l+1)\times N\times N\times C^{l-1}$. Therefore, the total reduction ratio of in the computation can be given as:
\begin{displaymath}
Reduction=\frac{(\frac{1}{2}\times K\times K\times C^l-1)\times N\times N\times C^{l-1}}{N\times N\times K\times K\times C^{l-1}\times C^l}=\frac{1}{2}-\frac{1}{K\times K\times C^l}
\end{displaymath}
Since in most cases we have $K\times K\times C^{l}\gg 2$, our approach has approximately 50\% FLOPs as compared to its plain counterpart. Thus, models equipped with comb convolution are typically faster than those equipped with standard convolution especially when applied to very width neural networks.

\subsection{Model capacity}

Generally, information flow from standard convolution is inefficient since too many connections have to be computed. Most prior studies, such as group-wise convolution\cite{31}, depth-wise convolution\cite{44}, and point-wise convolution\cite{33}, focused on accelerating convolution by constructing sparse connections. However, too much emphasis on sparse connections usually leads to the reduction of model parameters as well. Loss of parameters reduces dimension of the parameter searching space, resulting in the loss of model capacity, which leads to sharp decreasing in generalization performance. By contrast, comb convolution utilizes an architecture which reduces connections while still keeping the same parameter search space, and therefore has no loss of model capacity.

Specifically, the convolution operation can be represented as a matrix-vector multiplication operation. If the input and output are unrolled into vectors, the comb convolution will be represented as a sparser matrix than standard convolution\cite{34}. Without loss of generality, take for example convolving a 3$\times$3 kernel over a 4$\times$4 input using unit stride, the sparse matrix can be represented as:
\setcounter{MaxMatrixCols}{16}
\setlength{\arraycolsep}{2.2pt}
\[ \begin{matrix}
\begin{pmatrix} w_{0,0}&w_{0,1}&w_{0,2}&0&w_{1,0}&w_{1,1}&w_{1,2}&0&w_{2,0}&w_{2,1}&w_{2,2}&0&0&0&0&0\\
0&0&0&0&0&0&1&0&0&0&0&0&0&0&0&0\\
0&0&0&0&0&0&0&0&0&1&0&0&0&0&0&0\\
0&0&0&0&0&w_{0,0}&w_{0,1}&w_{0,2}&0&w_{1,0}&w_{1,1}&w_{1,2}&0&w_{2,0}&w_{2,1}&w_{2,2}
\end{pmatrix}
\end{matrix} \]
Note that the non-zero elements are the constant or the elements $w_{i,j}$ of the kernel, where $i$ and $j$ being the row and column of the kernel respectively.

\subsection{Point-wise feature reuse}
\label{I}

An effective model requires a sufficient number of convolution layers to capture good representations from input data. However, as the number of the convolution layers grows beyond a threshold, challenges such as vanishing gradient and other issues can arise. In order to make gradient information preservation, recent works, $e.g.$, ResNet-like and DenseNet-like, combine features through element-wise summation or concatenation before they are passed into subsequent layers, which is known as feature reuse. Comb convolution, in a similar spirit, exploit a kind of point-wise feature reuse, yielding models with high efficiency and representation capability.

Instead of using extra bypasses, we embed the cross-layer connections into convolutions to improve information flows and gradients throughout the network, that is, splitting the feature mappings of each layer into two branches, $i.e.$, convolution mapping and uniform mapping. On the one hand, convolution mapping branches propagate through weight layers and are mainly responsible for modeling spatial and channel feature maps. On the other hand, uniform mapping branches directly join the next layer without concerning any weight layers to ensure information flowing. Therefore, comb convolution efficiently models the dependencies in spatial dimensions while reduces computational complexity by half, which makes it possible to use more feature channels and larger network capacity. What’s more, since each layer directly preserves half the information of its preceding layer, the subsequent layer is freed from the burden of having to relearn previously useful features, thus alleviating issues with vanishing gradients. We also observe that when comb convolution operators are stacked, the effect of composited feature representation is profound.

\section{Experiments}

In this section, we evaluate the proposed comb convolution on the CIFAR-10 and CIFAR-100 benchmark classification datasets. Ablation studies compare performances and computational costs of various models configurations, which provide a useful guidance for network design. Extensive experiments are also conducted to confirm the effectiveness and efficiency of integrating our approach with VGGNets, Xception and SE-Net architectures respectively.

\subsection{Experimental setups}

\paragraph{Datasets}
The CIFAR datasets \cite{37}, CIFAR-10 and CIFAR-100, consist of   colored images. Both datasets contain 60,000 images, which are split into 50,000 training images and 10,000 test images. CIFAR-10 dataset has 10 classes, with 6,000 images per class. CIFAR-100 dataset is similar to CIFAR-10 dataset, except that has 100 classes, each of which contains 600 images. The standard data-augmentation scheme, in which the images are zero-padded with 4 pixels on each side, randomly cropped to produce   images, and horizontally mirrored with probability 0.5 are adopted in our experiments, according to usual practice \cite{39,40,41,42}.

\paragraph{Training settings}
All networks are trained using stochastic gradient descent (SGD) with a weight decay of $10^{-4}$ and momentum of 0.9. The parameters are initialized according to \cite{38}. On CIFAR-10 and CIFAR-100, we train for 300 epochs, with a mini-batch size of 100 on the NVIDIA 1080Ti GPU. The initial learning rate is set to 0.1 and is reduced by a factor 10 at the 50\%, 75\% of the training procedure. Unless otherwise specified, we adopt nonlinear activation right after each convolution operation. Batch Normalization (BN) is performed right after each nonlinear activation. We do not use any dropout regularization in this paper.

\begin{table}
\caption{Classification error results of comb convolutional neural networks with different model configurations on the CIFAR-10 dataset.}
\center
\scalebox{0.88}{
\begin{tabular}{c|c|c|c|c|c}
  \hline % horizontal line
 \multirow{1}{*}{\textbf{Models}} &\textbf{Depth}  &\textbf{Width}  &\textbf{BN strategy}   &\textbf{Mask interleaving} &\textbf{Acc.(\%)}\\
  \hline
  \hline
  \multirow{12}{*}{Comb CNNs}    &\multirow{6}{*}{8}    & 32     &\multirow{5}{*}{Pre-BN}    &$\surd$       &85.68\\
  \cline{3-3} \cline{5-6}
  &   & 48  &   & $\surd$  & 87.56\\
  \cline{3-3} \cline{5-6}
  &   & 64  &  & $\surd$  & 88.28\\ %first cell is occupied by the multirow
  \cline{3-3} \cline{5-6}%short partial horizontal lines from column 2 to column 5
  &  & 96  &   & $\surd$  & 89.74\\ %first cell is occupied by the multirow
  \cline{3-3} \cline{5-6}%short partial horizontal lines from column 2 to column 5
  &   & 64  & &   & 87.86\\ %first cell is occupied by the multirow
  \cline{3-6}
  &  & 64  & \multirow{1}{*}{Post-BN} & $\surd$  & 88.46\\
  \cline{2-6}
  &\multirow{6}{*}{16} & 32 & \multirow{5}{*}{Pre-BN}  &$\surd$   &88.36\\ %end line
  \cline{3-3} \cline{5-6}%short partial horizontal lines from column 2 to column 5
  &  & 48  &  & $\surd$  & 89.55\\ %first cell is occupied by the multirow
  \cline{3-3} \cline{5-6}%short partial horizontal lines from column 2 to column 5
  &  & 64  &  & $\surd$  & 90.53\\ %first cell is occupied by the multirow
  \cline{3-3} \cline{5-6}%short partial horizontal lines from column 2 to column 5
  &  & 96  &  & $\surd$  & 91.50\\ %first cell is occupied by the multirow
  \cline{3-3} \cline{5-6}%short partial horizontal lines from column 2 to column 5
  &  & 64  &  &   & 90.00\\ %first cell is occupied by the multirow
  \cline{3-6}
  &  & 64  &  \multirow{1}{*}{Post-BN} & $\surd$  & 90.42\\
  \hline
\end{tabular}}
\label{table1}
\end{table}

\begin{figure}
\centering
\subfloat[]{\label{fig2_a}\includegraphics[width=1\columnwidth]{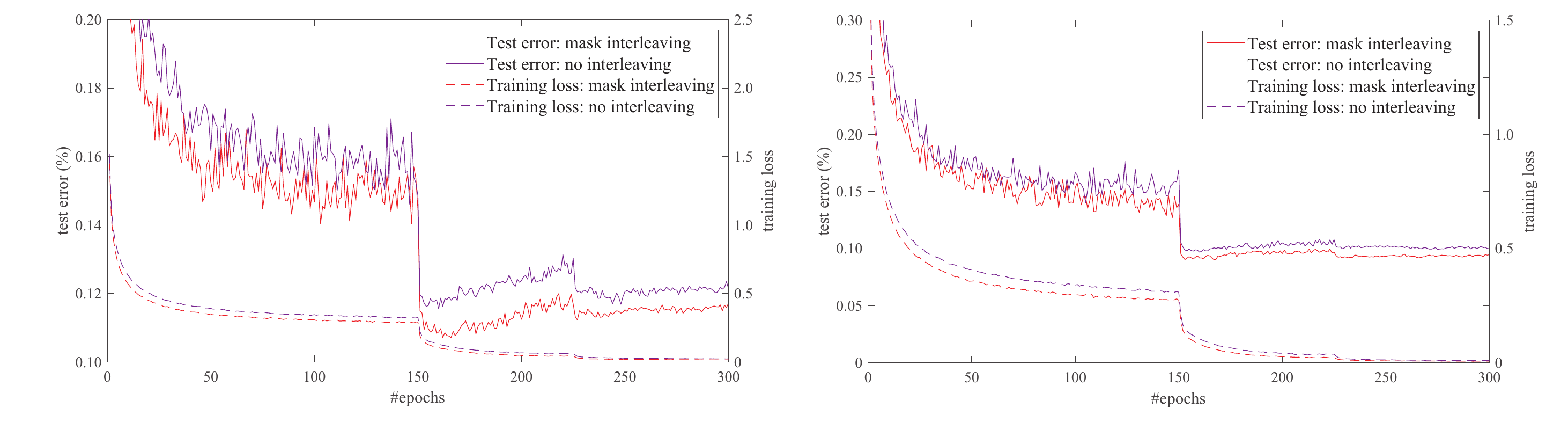}}\
\subfloat[]{\label{fig2_b}\includegraphics[width=1\columnwidth]{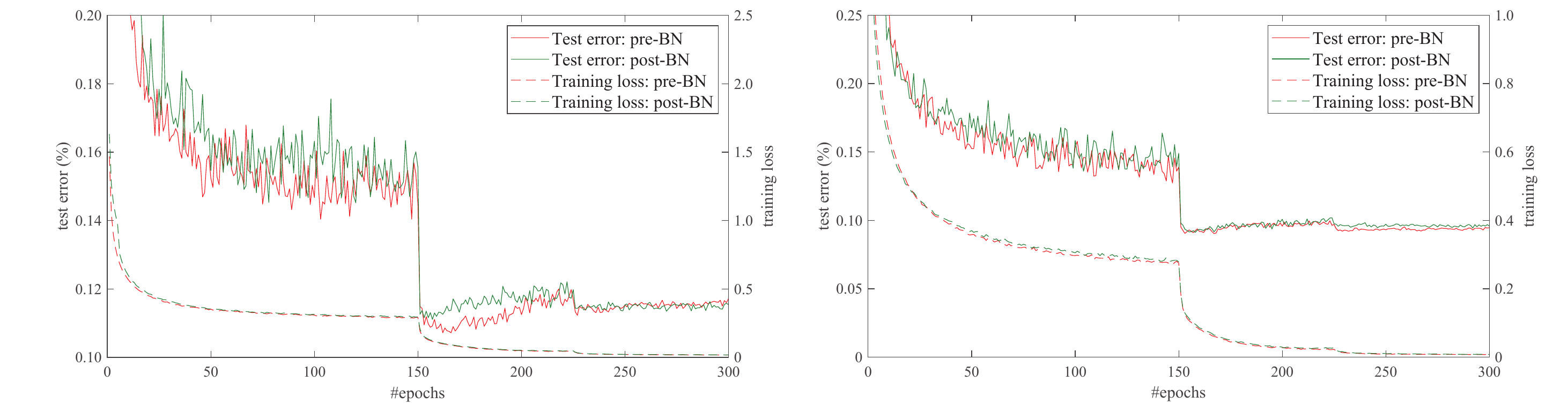}}\
\subfloat[]{\label{fig2_c}\includegraphics[width=1\columnwidth]{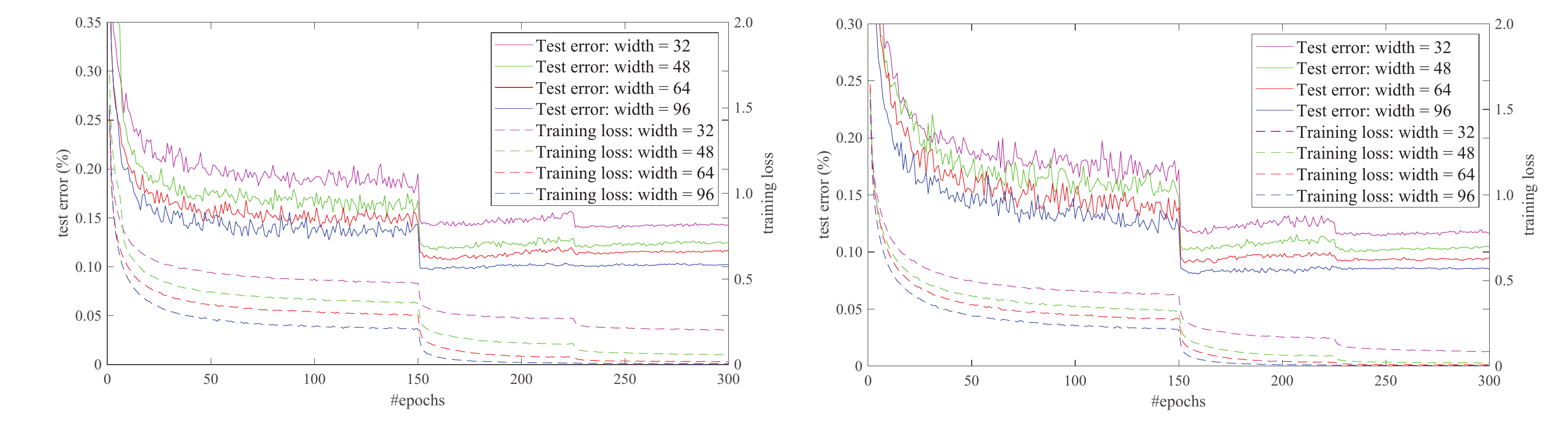}}\
\caption{Illustrations of the training and testing curves of the 8-layer networks ($top$) and 16-layer networks ($below$) with comb convolution on CIFAR-10. Solid curves denote test error, and dotted curves denote training loss. (a) The effect of mask interleaving operation. (b) Comparison of different BN strategy. (c) Network width influence.} \label{fig2}
\end{figure}

\subsection{Ablation studies on CIFAR-10}

Our ablation experiments are performed on CIFAR-10 classification dataset. By using a simply stack of comb convolution operators, we investigate the effect of 1) mask interleaving operation, 2) pre-BN strategy, and 3) network width. The results are presented in Table ~\ref{table1}.

\paragraph{The effect of mask interleaving operation}
Considering uniform mappings from a certain layer only learn the cross-channel correlations. This property will block information flow between spatial dimension and weaken cross-spatial representation. Hence, we proposed a fixed mapping permutation, $i.e.$, mask interleaving, which enables uniform mapping to simultaneously model dependencies in both the spatial and channel dimensions. The evaluations are performed under two different scales. Figure ~\ref{fig2_a} depicts the training and testing curves of the mask interleaving vs. no interleaving. It is clear that mask interleaving consistently boosts classification accuracy for different depth, showing the efficacy of cross-spatial information interchange. Therefore, employing mask interleaving is efficient strategy to further improve the capacity of the information representation.

%\begin{figure}
%\centering
%\subfloat[]{\includegraphics[width=0.3\columnwidth]{mask_2.pdf}}\
%\subfloat[]{\includegraphics[width=0.3\columnwidth]{bn_1.pdf}}\
%\subfloat[]{\includegraphics[width=0.3\columnwidth]{width_2.pdf}}\
%\caption{The test error of networks with comb convolution at different width on the
%CIFAR-10 benchmark.} \label{bn}
%\end{figure}

\begin{figure}
\centering
{\includegraphics[width=0.65\columnwidth]{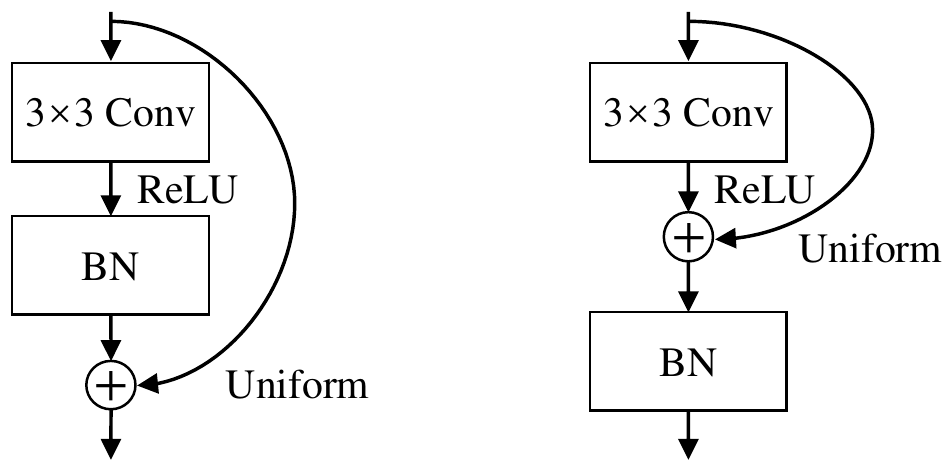}}\
\caption{The schema of the pre-BN strategy($left$) and the post-BN strategy($right$).} \label{fig3}
\end{figure}

\paragraph{Pre-BN strategy}
Batch Normalization (BN)\cite{36} is obviously essential in facilitating convergence and improving performance. However, at training time, BN requires multiple multiplication, $i.e.$, dividing by the running variance. As a result, the number of scaling calculations is quite large especially in the case of convolutional neural network. To be efficiently compatible with BN, we conduct experiments on comb convolution with different structures, $i.e.$, BN before addition (denoted as pre-BN) and BN after the addition (denoted as post-BN). Note that the pre-BN structure is applied only to the convolutional layer rather than the all layers. Figure ~\ref{fig3} shows an example, where both designs have similar space complexity. Experimental results (see Table ~\ref{table1} and Figure ~\ref{fig2_b}) indicate that the pre-BN strategy is more efficient (lesser FLOPs with similar accuracy) as compared to post-BN.

\paragraph{Network width}
We compared the performance at different widths and reported the results in Table ~\ref{table1} and Figure ~\ref{fig2_c}. By exploring different scales of models, we can observe that increasing the network width can consistently contribute to performance gain. For example, the classification accuracy of the 16-layer network equipped with 32-channel comb convolution is 88.36\%, while that of the 48-channel reaches 89.55\%, that of the 64-channel reaches 90.53\%, and that of the 96-channel reaches 91.50\%. Since the efficient design of comb convolution, we can use more channels for given computation budget, thus usually resulting in better performance; this indicates the great potential of comb convolution in applications on low-end devices.

\subsection{Integrated models on CIFAR-100}

We choose the representative multi-channel, single-channel and channel attention CNNs as baseline models, $i.e.$, VGGNets, Xception, and SE-Net. For fair comparisons, we simply replace the corresponding standard convolution with the proposed comb convolution, and keep most of the same model configurations as described in \cite{21,12,22}. The detailed model configuration is as follows:

\paragraph{VGGNets}
Following the design principle of VGGNets, we use $3\times 3$ comb convolution as the basic convolution operator. Different from \cite{21}, we add a BN layer after each of the standard convolutional layer for faster converge.

\paragraph{Xception}
The core idea of Xception lies in the depthwise separable convolution, which generalized the factorization idea and decomposed the convolution into a depthwise convolution and a $1\times 1$ convolution. We replaced the depthwise convolution in the Xception with the proposed single-channel comb convolution. We also modified the convolution strides of the initial layer and the exit flow block to match the CIFAR dataset.

\paragraph{SE-Net}
An SE block adaptively re-calibrates channel-wise feature responses by explicitly modeling interdependencies between channels. The proposed comb convolutions focus on spatial dimension that is orthogonal to this method. We integrate SE blocks into the middle flow block of Xception, in a similar schema to the SE-ResNet \cite{11}.

We use exactly the same settings to train these models. Results are showed in Tables ~\ref{table2}, and Table ~\ref{table3}. Although the baseline is slightly better than our approach as shown in Table ~\ref{table2}. It is, however, worth noting that the FLOPs of comb convolution is only $\frac{1}{2}$ relative to that of standard convolution and this ratio will decrease when BN is added. This property can be very useful when the memory constraint is tight. As shown in Table ~\ref{table3}, our models Comb-Xception has better accuracy as compare to the baseline. Notably, adding SE block improves accuracy from 69.67\% to 69.70\%, which proves comb convolution is orthogonal and complementary to the channel-wise method.

\begin{table}
\caption{The test accuracy for the baseline architectures VGGNets and their respective comb convolution counterparts on CIFAR-100.}
\center
\begin{tabular}{c|ccc}
  \hline
  \multirow{1}{*}{\textbf{Models}} &\textbf{Depth}   &\textbf{MFLOPs}    &\textbf{Acc.(\%)}\\
  \hline
  \hline
  \multirow{4}{*}{VGGNets} & 11  & 169.80 &63.79\\ %end line
  \cline{2-4} %short partial horizontal lines from column 2 to column 5
  &13   & 245.30 & 68.45\\ %first cell is occupied by the multirow
  \cline{2-4} %short partial horizontal lines from column 2 to column 5
  &16   & 330.24  & 70.30\\ %first cell is occupied by the multirow
  \cline{2-4} %short partial horizontal lines from column 2 to column 5
  &19   & 500.11 & 69.42\\ %first cell is occupied by the multirow
  \hline
  \multirow{4}{*}{Comb-VGGNets} &11    & 93.50  & 57.82\\
  \cline{2-4} %short partial horizontal lines from column 2 to column 5
  &13   & 131.24 & 62.55\\ %first cell is occupied by the multirow
  \cline{2-4} %short partial horizontal lines from column 2 to column 5
  &16   &  173.71 & 65.63\\ %first cell is occupied by the multirow
  \cline{2-4} %short partial horizontal lines from column 2 to column 5
  &19   & 258.65  & 63.33\\ %first cell is occupied by the multirow
  \hline
\end{tabular}
\label{table2}
\end{table}

\begin{table}
\caption{The test accuracy for the baseline architectures Xception and SE-Xception, and their respective comb convolution counterparts on CIFAR-100.}
\center
\begin{tabular}{c|cc}
  \hline
  \multirow{1}{*}{\textbf{Models}}  &\textbf{MFLOPs}    &\textbf{Acc.(\%)}\\
  \hline
  \hline
  \multirow{1}{*}{Xception} & 208.85 &69.53\\
  \multirow{1}{*}{Comb-Xception}& 206.71 &69.67\\
  \hline
  \multirow{1}{*}{SE-Xception} & 209.91 &69.89\\
  \multirow{1}{*}{Comb-SE-Xception} & 207.77 &69.70\\
  \hline
\end{tabular}
\label{table3}
\end{table}

%\begin{figure}
%\centering
%{\includegraphics[width=0.9\columnwidth]{vgg.pdf}}\
%\caption{Reduction in computations on VGGNets with regard to different depths.} \label{fig4}
%\end{figure}

\subsection{Visualization analysis}
\label{II}

\begin{figure}
\centering
\subfloat[]{\label{fig5_a}\includegraphics[width=0.98\columnwidth]{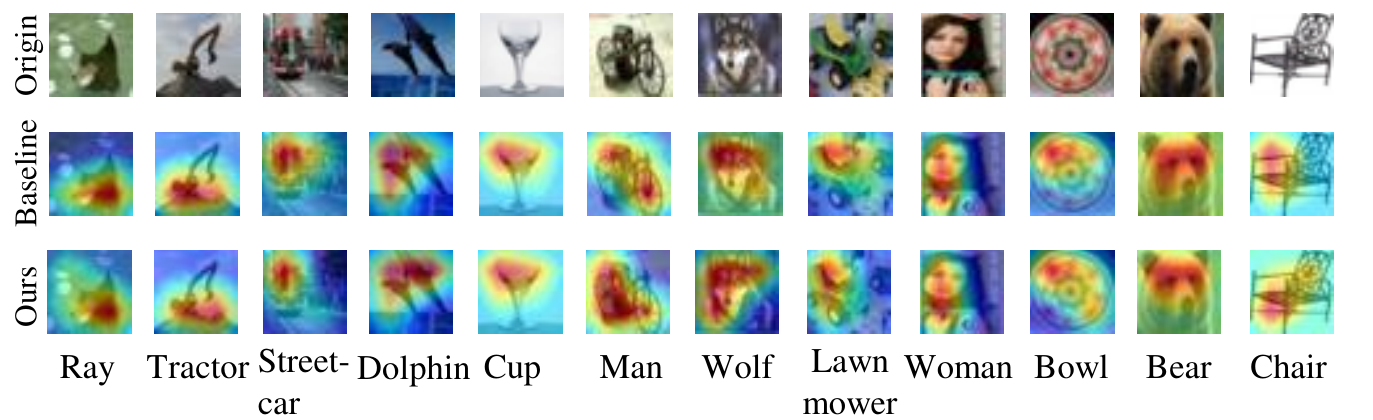}}\
%\hspace{0.01cm}
\subfloat[]{\label{fig5_b}\includegraphics[width=0.98\columnwidth]{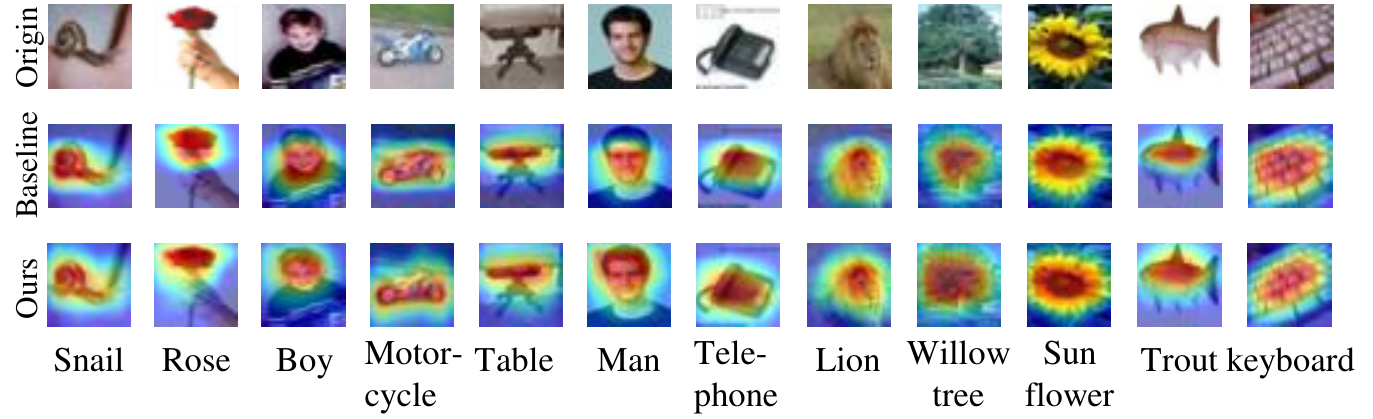}}\
\caption{Visualization of the class activate mapping (CAM) on different object sizes. (a) Visualization of CAM in the last convolutional layer of VGG-16 ($middle$) and Comb-VGG-16 ($below$). (b) Visualization of CAM in the penultimate convolutional layer of Xception ($middle$) and Comb-Xception ($below$). As expected, models equipped with comb convolution have similar activation regions compared to their plain counterparts. Figure best viewed in color.} \label{fig5}
\end{figure}

Models equipped with comb convolution can capture the same receptive field as standard convolution so as to model dependencies in both spatial and channel dimensions with high efficiency. To understand this, we visualize the class activate mapping (CAM) using Grad-CAM \cite{35}, which is the class-discriminative localization technique for visual explanations. As shown in Figure ~\ref{fig5}, stronger CAM regions are covered by lighter colors. For instance, as we visualize final convolutional layer of the VGG-16 equipped with comb convolution and the baseline respectively, we see both convolution operators have similar activation regions for objects of all size. Notably, the inference holds true for the case where employing the single-channel comb convolution as well, as shown in Figure ~\ref{fig5_b}. Therefore, such ability of efficiently and effectiveness spatial sampling makes comb convolution be potentially valuable for object region mining tasks.

\section{Conclusion}

In this work, we focused on reducing spatial redundancy that widely exists in standard convolution and proposed a novel convolution operator, namely comb convolution, for designing efficient convolutional architecture. By inserting uniform mappings into standard convolutions, comb convolution eliminates nearly 50\% of connections and achieves 2$\times$ FLOPs based speed boost as compared to the standard convolution. Further, comb convolution exploits a new kind of feature reuse, namely point-wise feature reuse, which helps strike an excellent trade-off between representation capability and computational cost. Notably, comb convolution is orthogonal to existing state-of-the-art methods that utilize channel-wise operations, thus can be integrated with these architectures to further improve efficiency. Extensive experiments are performed on CIFAR-10 and CIFAR-100 benchmark classification datasets, which confirm the effectiveness and efficiency of our approach with VGGNets, Xception and SE-Net architectures.

\bibliographystyle{IEEETrans}
\bibliography{ref}

\end{document}